\documentclass[11pt]{article}

\usepackage[]{acl}

\usepackage{times}
\usepackage{latexsym}

\usepackage[T1]{fontenc}
\usepackage[utf8]{inputenc}

\usepackage{microtype}

\usepackage{inconsolata}

\usepackage{graphicx}
\usepackage{booktabs}
\usepackage{multirow}
\usepackage{paralist}
\title{Samasāmayik: A Parallel Dataset for Hindi-Sanskrit Machine Translation}

\author{
 \textbf{N J Karthika\textsuperscript{1}},
 \textbf{Keerthana Suryanarayanan\thanks{Work done while pursuing internship at IIT Roorkee}\textsuperscript{2}},
 \textbf{Jahanvi Purohit\textsuperscript{1}},
 \\
 \textbf{Ganesh Ramakrishnan\textsuperscript{1}},
 \textbf{Jitin Singla\textsuperscript{3}},
 \textbf{Anil Kumar Gourishetty\textsuperscript{3}},
 \\
 \textsuperscript{1}Indian Institute of Technology Bombay,
 \textsuperscript{2}Geakminds Technologies Private Limited,
 \\
 \textsuperscript{3}Indian Institute of Technology Roorkee
 \\
 }
\begin{document}
\maketitle
\begin{abstract}
We release Samasāmayik, a novel, meticulously curated, large-scale Hindi-Sanskrit corpus, comprising 92,196 parallel sentences. Unlike most data available in Sanskrit, which focuses on classical era text and poetry, this corpus aggregates data from diverse sources covering contemporary materials, including spoken tutorials, children's magazines, radio conversations, and instruction materials. We benchmark this new dataset by fine-tuning three complementary models - ByT5, NLLB and IndicTrans-v2, to demonstrate its utility. Our experiments demonstrate that models trained on the Samasamayik corpus achieve significant performance gains on in-domain test data, while achieving comparable performance on other widely used test sets, establishing a strong new performance baseline for contemporary Hindi-Sanskrit translation. Furthermore, a comparative analysis against existing corpora reveals minimal semantic and lexical overlap, confirming the novelty and non-redundancy of our dataset as a robust new resource for low-resource Indic language MT.
 % \\ \newline \Keywords{parallel corpus, machine translation, Sanskrit, Hindi} 
 
\end{abstract}
\section{Introduction}
Sanskrit is one of the oldest and most systematic languages in the world. Sanskrit not only serves as the linguistic foundation for many other languages, but is also a source of a vast knowledge base in various fields, including science, literature, philosophy, etc. From the grammatical point of view, Sanskrit is one of the most structured languages with well-defined rules by scholars like Pāṇini and his successors. Apart from being one of the oldest languages, Sanskrit's logical structure and clarity make it relevant even today in modern linguistic studies, especially in computational linguistics. Given this rich tradition and the well-structured grammar, Natural Language Processing (NLP) for the Sanskrit language is of great significance. Compared to the vast amount of documents and literature available in Sanskrit, there is considerably less digitized data available for Sanskrit, making it a low-resource language. Among the available digital content in Sanskrit, the main focus is on classical era texts and poetry. In addition to the challenge of limited digital data availability, there is further difficulty in obtaining parallel sentences, which is imperative for tasks like Machine Translation (MT).

Existing MT datasets involving Sanskrit include English-Sanskrit parallel dataset contributions from \citet{maheshwari-etal-2024-samayik, aralikatte-etal-2021-itihasa, punia-etal-2020-improving} etc. Many of these contributions consist of verses from epics such as Ramāyana, Mahābhārata etc, poems and other literature from the classical era. The availability of digital data for Sanskrit in the context of contemporary prose remains limited. Bharat Parallel Corpus Collection (BPCC) \cite{gala2023indictrans2} is another comprehensive, large-scale English-Indic parallel dataset, which involves 22 languages. 

\begin{figure*}[!htbp]
    \centering
    \includegraphics[width=0.95\textwidth]{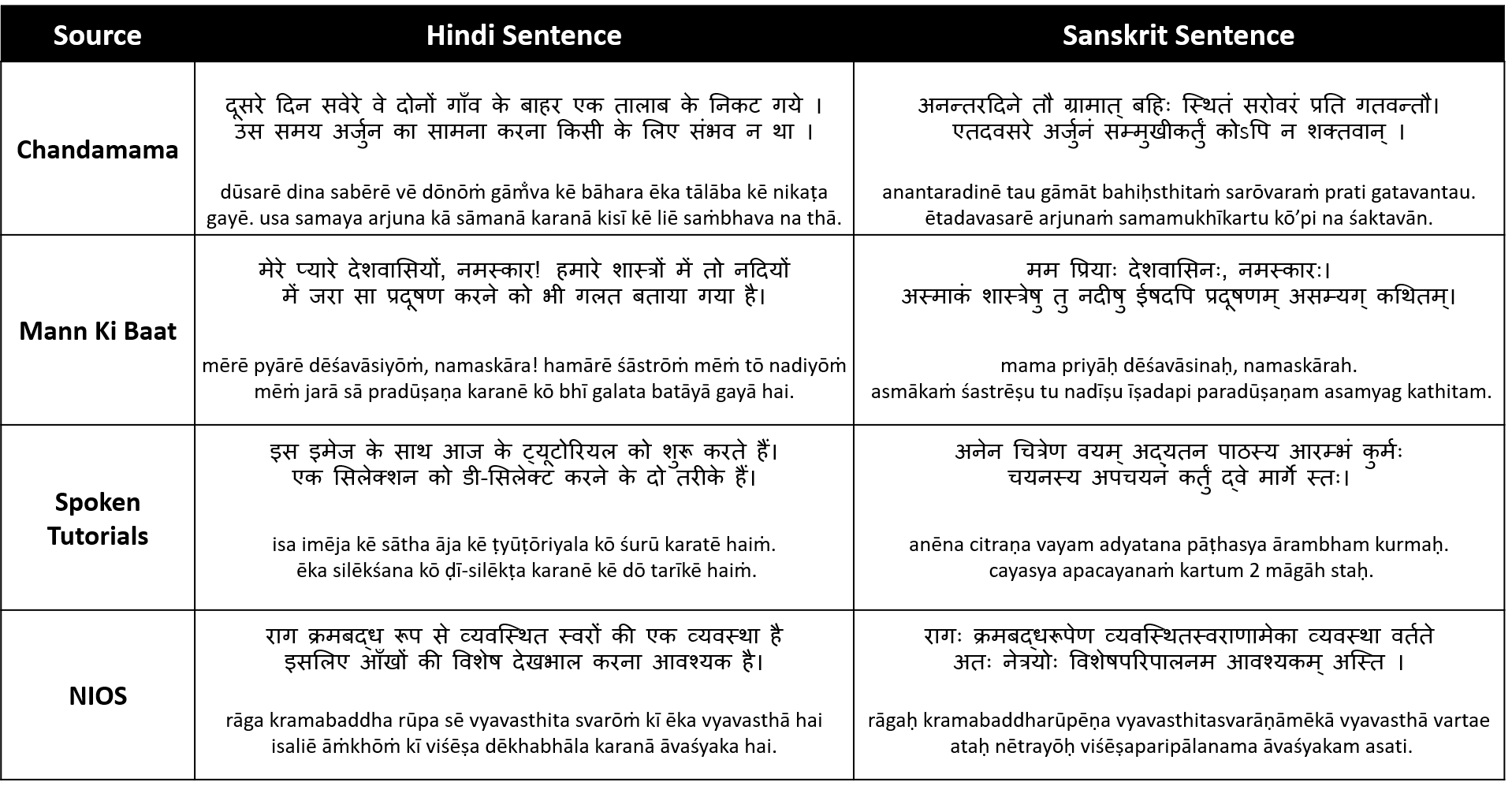}
    \caption{Example sentences of the dataset from various sources. }
    \label{fig:examples}
\end{figure*}

To address this gap, we introduce Samasāmayik, a new large-scale Sanskrit-Hindi parallel corpus, consisting of 92,196 sentence pairs (90,016 pairs for training and 1,839 pairs for testing). The dataset is curated from diverse contemporary sources, including spoken tutorials, children's magazine (Chandamama), radio conversation (Mann ki Baat) and instruction materials. Details of these data sources, the process of sentence alignment etc. are given in Section~\ref{sec:data}. Figure~\ref{fig:examples} show some of the example parallel sentences from the proposed dataset. Detailed data statistics are shown in Tables \ref{tab:dataset-stats-train} and \ref{tab:dataset-stats-test}.

For comparative evaluation, we additionally use the BPCC dataset. Since many English sentences are shared across the English-Indic language pairs, it's possible to extract Indic-Indic data from this corpus via pivoting through English. We extracted 79,977 Hindi-Sanskrit parallel sentences from BPCC, and use it solely as a baseline dataset to evaluate and contrast the quality and utility of Samasāmayik. These BPCC-derived pairs are not part of the proposed dataset.

To assess the quality and utility of the proposed Samasāmayik dataset, we use three complementary models - ByT5 \cite{xue2022byt5} (a byte-level text-to-text model) , NLLB-1.3B parameter model \cite{costa2022nllb} (a multilingual translation model), and IndicTransV2 \cite{gala2023indictrans2} (a transformer based translation model, specific to Indian languages). We fine-tune these models on the proposed training data and compare the performance against their fine-tuning on existing BPCC data. Evaluation of the models is performed using the proposed test data and two other widely-accepted translation evaluation test sets - IN22\cite{gala2023indictrans2} and Flores-200\cite{costa2022nllb}. The details of the model fine-tuning, evaluation datasets and metrics are provided in Section~\ref{sec:expnres}.

Following are our contributions:
\begin{asparaenum}
    \item We propose a novel, meticulously curated, sentence aligned large scale Hindi-Sanskrit parallel data, for contemporary MT. We release the dataset and code to support future research and reproducibility.\footnote{https://github.com/karthika95/samasaamayik}
    \item We verify the utility of the dataset by training three complementary models for the task of MT and measuring their performance across three different test-sets.
    \item We also verify the novelty of our dataset by comparing the data with an existing dataset using various similarity metrics. The results show that our dataset is significantly distinct, with minimal semantic and lexical overlap with the existing data.
\end{asparaenum}

% \section{Related Works}

\section{Dataset Details}
\label{sec:data}
In this section, we provide details about the various sources of our data, and the processes of translation and alignment wherever applicable. Our dataset is a parallel sentence corpus of Hindi-Sanskrit pairs, comprising 92,196 sentence-aligned data points from 4 distinct sources \emph{viz.,} Chandamama, Mann Ki Baat (MKB), Spoken Tutorials and National Institute of Open Schooling (NIOS) data. The dataset aims to accommodate translation pairs from modern Hindi and Sanskrit prose. 

\subsection{Chandamama}
Chandamama\footnote{https://chandamama.page/} is an Indian monthly magazine for children. It is known for its long-running mythological stories. It was founded in July 1947, and the last edition was July 2013. It was published in 13 languages, including English, Hindi, and Sanskrit. We use the Hindi and Sanskrit versions of 9 issues, where each issue contains stories with titles. Optical Character Recognition (OCR) was performed on the PDF documents of the magazines to extract the text. Our team of language experts, conversant in both Hindi and Sanskrit, performed sentence alignment from the respective text files. The mapping in this case was simpler as there were direct sentence-by-sentence correspondences between the Hindi and Sanskrit sentences. This set includes 8,870 parallel sentences.

\subsection{Mann Ki Baat (MKB)}
Mann Ki Baat is a monthly radio program hosted by the Prime Minister of India, Narendra Modi. It was first broadcast in October 2014 and serves as a platform for the Prime Minister to address the nation on social, cultural, and contemporary issues, share inspiring stories of ordinary citizens, and discuss the government's initiatives and policies. Sanskrit translations of 25 older episodes, done by professionals but unofficial, are available in the public domain. Sanskrit translations of 11 more recent episodes were done by in-house experts. We use these expert translations to manually align the Sanskrit sentences with the official Hindi transcripts\footnote{https://www.narendramodi.in/hi/mann-ki-baat} throughout the 36 episodes. The MKB set contains 6424 Hindi-Sanskrit parallel sentences.

\subsection{Spoken Tutorials}
The Spoken Tutorial project offers a huge collection of video lessons for teaching students how to utilize open-source software. The tutorials are written by domain experts, which are then translated into many languages by skilled translators. We scraped videos and transcripts with Hindi and Sanskrit translations from their website \footnote{https://spoken-tutorial.org/}. We retrieved transcripts from 254 videos, each lasting an average of 10 minutes. Transcripts are manually prepared. Five language experts aligned each transcript's Hindi sentences with their corresponding Sanskrit sentences. Experts align and combine Hindi and Sanskrit sentences in transcripts where a one-to-one connection is non-existent. The final set contains about 24,930 sentences.

\subsection{NIOS}
The National Institute of Open Schooling (NIOS)\footnote{https://www.nios.ac.in/} is an autonomous organization under the Ministry of Education, Government of India. It provides flexible education to students who are unable to attend regular schools, offering courses at secondary and senior secondary levels, vocational programs, and continuing education opportunities for adults and professionals. NIOS offers self-learning materials like textbooks and study guides, designed for independent curriculum understanding. We obtained the study materials of the NIOS' Indian knowledge tradition courses, which are available in Hindi and Sanskrit. Each course comprises multiple topics presented as PDF files. We utilize PDF parsers to convert the PDF content into text format while retaining all the information. Our team of five linguistic experts proficient in Hindi and Sanskrit aligned the sentences from the respective text files.

\begin{table}[t]
\centering
\small
\scalebox{0.85}{
    \begin{tabular}{lcc}
    \toprule
     & \textbf{Hindi} & \textbf{Sanskrit} \\
    \midrule
    Total Sentences           & 90,016  & 90,016 \\
    Total Words              & 1,167,555 & 800,704 \\
    Unique Words           & 97,284  & 180,754 \\
    Type--Token Ratio (TTR)   & 0.083  & 0.226 \\
    Average Sentence Length (words) & 12.97 & 8.90 \\
    Median Sentence Length (words)  & 11.00 & 8.00 \\
    Max Sentence Length (words)     & 285   & 167 \\
    Average Sentence Length (chars)  & 68.15 & 67.59 \\
    Max Sentence Length (chars)      & 1,600 & 1,500 \\
    Average Word Length (chars)      & 4.32  & 6.69 \\
    \bottomrule
    \end{tabular}
    }
\caption{Dataset statistics of Hindi and Sanskrit Training data}
\label{tab:dataset-stats-train}
\end{table}

\subsection{Combined Data}
Combining all the above mentioned data sources after filtering out some empty entries, we propose a Hindi-Sanskrit parallel corpus, consisting of 90,016 sentences as part of training data and 1,839 sentences for testing. The detailed dataset statistics of the training and test data are mentioned in tables \ref{tab:dataset-stats-train} and \ref{tab:dataset-stats-test}. Figure~\ref{fig:examples} show some of the example parallel sentences from the proposed dataset.

\begin{table}[t]
\centering
\small
\scalebox{0.85}{
    \begin{tabular}{lcc}
    \toprule
    \textbf{Statistic} & \textbf{Hindi} & \textbf{Sanskrit} \\
    \midrule
    Total Sentences           & 1,839  & 1,839 \\
    Total Words              & 23,640 & 16,251 \\
    Unique Words           & 7,199  & 9166 \\
    Type--Token Ratio (TTR)   & 0.305  & 0.564 \\
    Average Sentence Length (words) & 12.85 & 8.84 \\
    Median Sentence Length (words)  & 11.00 & 8.00 \\
    Max Sentence Length (words)     & 75   & 42 \\
    Average Sentence Length (chars)  & 67.83 & 67.28 \\
    Max Sentence Length (chars)      & 379 & 330 \\
    Average Word Length (chars)      & 4.35  & 6.70 \\
    \bottomrule
    \end{tabular}
    }
\caption{Dataset statistics of Hindi and Sanskrit Test data}
\label{tab:dataset-stats-test}
\end{table}

\section{Experiments and Results}
\label{sec:expnres}
% \subsection{Experimental Setup}
We perform various experiments to verify the quality, usability and novelty of our proposed dataset. We use three complementary models: ByT5-base \cite{xue2022byt5}, a multilingual byte-level text-to-text transformer model, NLLB-200 (1.3B parameter model) \cite{costa2022nllb}, a multilingual translation task specific model and IndicTransV2 \cite{gala2023indictrans2} (a transformer based translation model, specific to Indian languages). We finetune these models to perform translation by training them on the proposed corpus's training data and also with an existing Hindi-Sanskrit parallel data, extracted from the BPCC corpus. The translation performance of the finetuned models are then tested on three different test sets described in Section~\ref{sec:evaldata}. The details of the finetuning process, including the training configurations, hyperparameters, and the model architecture, are given in Tables \ref{tab:byt5_model_card} and \ref{tab:nllb_model_card} in Section~\ref{sec:supplement}.

To verify the novelty of the proposed dataset, we perform some comparative analysis experiments with respect to an existing publicly available vast, parallel dataset Bharat Parallel Corpus Collection(BPCC), in terms of the sentence overlap ratio and similarity metrics. We randomly sampled 10,000 sentences from each corpus and computed the Cosine Similarity (to find Semantic overlap) and Jaccard similarity (to find lexical overlap). The results and the observation are as follows:
\begin{asparaenum}
\item \textbf{\textit{Cosine Similarity}}\\
We measured the pairwise cosine similarity between the sampled sentences from both the corpora using LaBSE sentence embeddings\cite{feng-etal-2022-language}. The number of samples with a cosine similarity > 0.9 was 0, indicating that none of the sentences in the sample set are even near-duplicates of the sentences in the BPCC data. This result also suggests that our dataset is non-redundant and not mined or duplicated from the existing resource.\\
The average maximum cosine similarity across the sample was 0.515, implying that the sentences of our corpus are semantically distant from the existing corpus. This also supports the novelty and diversity of the content of the dataset.
\item \textbf{\textit{Jaccard Similarity}}\\
For each sentence in the sample from our dataset, we computed the Jaccard Similarity with all the sentences from that of BPCC. On an average, the most similar sentence from BPCC has only about 17\% token-overlap with the corresponding sentence in our dataset. This measure is also indicative that the proposed dataset has a broader lexical coverage as compared to BPCC.
\end{asparaenum}

Summarily, the above experiment results indicate that the proposed Hindi $\leftrightarrow$ Sanskrit parallel dataset is novel, non-redundant, semantically and lexically diverse.

\subsection{Experimental Setup}
To assess the quality and utility of the proposed Samasāmayik Hindi$\leftrightarrow$Sanskrit parallel dataset, we use three complementary models - ByT5 \cite{xue2022byt5} (a byte-level text-to-text model) , NLLB-1.3B parameter model \cite{costa2022nllb} (a multilingual translation model), and IndicTransV2 \cite{gala2023indictrans2} (a transformer based translation model, specific to Indian languages). We fine-tune these models on the proposed training data and compare the performance against their training on existing BPCC data.
The details of the multilingual model architectures, their finetuning process, including the training configurations, and hyperparameters are given in Tables \ref{tab:byt5_model_card}, \ref{tab:indictrans_model_card} and \ref{tab:nllb_model_card}.

\paragraph{ByT5} 
\cite{xue2022byt5} is a byte-oriented sequence-to-sequence transformer model that processes raw text at the byte level, enabling language-agnostic handling of diverse scripts without relying on tokenization. It is a token-free extension of T5 and mT5 transformer architectures.

\label{sec:supplement}
\begin{table}[h]
\centering
\small
\scalebox{0.85}{
\begin{tabular}{ll}
\hline
\textbf{Model} & \textbf{\textit{google/byt5-base}} \\
\textbf{Architecture} & Encoder--Decoder (Seq2Seq) \\
\textbf{Parameters} & $\sim$580M \\
\textbf{Tokenizer} & Byte-level (UTF-8) \\
% \textbf{Languages} & Hindi (hi) $\rightarrow$ Sanskrit (sa) \\
% \textbf{Data Split} & 95\% train / 5\% val (seed=42) \\
\textbf{Max Lengths} & 1024 (hin), 512 (san) \\
\textbf{Batch Size} & 8 \\
\textbf{Optimizer} & AdamW \\
\textbf{Learning Rate} & 5e--5 (linear decay) \\
\textbf{Epochs} & 3 \\
% \textbf{Checkpointing} & Best loss retained per epoch \\
\textbf{Hardware} & 2 $\times$ NVIDIA A6000 GPU \\
\hline
\end{tabular}
}
\caption{Model card for \textbf{ByT5-Base} fine-tuned for Hindi$\leftrightarrow$Sanskrit translation.}
\label{tab:byt5_model_card}
\end{table}

\paragraph{NLLB}
\cite{costa2022nllb} is a large-scale multilingual translation system (an encoder-decoder transformer architecture), trained on 200 languages. It was designed to improve translation quality for both high and low-resource languages.

\begin{table}[h]
\centering
\small
\scalebox{0.81}{
\begin{tabular}{ll}
\hline
\textbf{Model} & \textbf{\textit{NLLB-200 Distilled 1.3B}} \\
\textbf{Base Architecture} & Transformer Encoder-Decoder (Seq2Seq) \\
\textbf{Parameters} & $\sim$1.29B \\
\textbf{Tokenizer} & SentencePiece (BPE, 256K vocab) \\
% \textbf{Languages} & Hindi (hi) $\rightarrow$ Sanskrit (sa) \\
% \textbf{Data Split} & 95\% train / 5\% val (seed=42) \\
\textbf{Max Lengths} & 256 tokens \\
\textbf{Batch Size} & 8 \\
\textbf{Optimizer} & AdamW \\
\textbf{Learning Rate} & 3e--5 (linear decay) \\
\textbf{Epochs} & 3 \\
% \textbf{Checkpointing} & Best loss retained per epoch \\
\textbf{Hardware} & 4 $\times$ NVIDIA A6000 GPU \\
\hline
\end{tabular}
}
\caption{Model card for \textbf{NLLB-200 (Distilled 1.3B)} fine-tuned for Hindi$\leftrightarrow$Sanskrit translation.}
\label{tab:nllb_model_card}
\end{table}

\paragraph{IndicTrans}
\cite{gala2023indictrans2} is a transformer-based encoder-decoder architecture, to cater to machine translation focusing on Indian languages.

\label{sec:supplement}
\begin{table}[h]
\centering
\small
\scalebox{0.85}{
\begin{tabular}{ll}
\hline
\textbf{Model} & \textbf{\textit{indictrans2-indic-indic-1B}} \\
\textbf{Architecture} & Transformer Encoder--Decoder (Seq2Seq) \\
\textbf{Parameters} & $\sim$1B \\
\textbf{Tokenizer} & SentencePiece (ULM, 256k vocab) \\
% \textbf{Languages} & Hindi (hi) $\rightarrow$ Sanskrit (sa) \\
% \textbf{Data Split} & 95\% train / 5\% val (seed=42) \\
\textbf{Max Lengths} & 128 tokens \\
\textbf{Batch Size} & 8 \\
\textbf{Optimizer} & AdamW \\
\textbf{Learning Rate} & 2e--5 (linear decay) \\
\textbf{Epochs} & 3 \\
% \textbf{Checkpointing} & Best loss retained per epoch \\
\textbf{Hardware} & 4 $\times$ NVIDIA A6000 GPU \\
\hline
\end{tabular}
}
\caption{Model card for \textbf{indictrans2-indic-indic-1B} fine-tuned for Hindi$\leftrightarrow$Sanskrit translation.}
\label{tab:indictrans_model_card}
\end{table}

\begin{table*}[!htbp]
\centering
\small
\begin{tabular}{llcccccc}
\toprule
\multirow{2}{*}{\textbf{Model}} & 
\multirow{2}{*}{\textbf{Training Data}} &
\multicolumn{3}{c}{\textbf{Hindi $\rightarrow$ Sanskrit}} & 
\multicolumn{3}{c}{\textbf{Sanskrit $\rightarrow$ Hindi}} \\
\cmidrule(lr){3-5} \cmidrule(lr){6-8}
 & & \textbf{BLEU} & \textbf{chrF++} & \textbf{WER} & \textbf{BLEU} & \textbf{chrF++} & \textbf{WER} \\
\midrule
\multicolumn{8}{c}{\textbf{Samasāmayik (Test Set)}} \\
\midrule
% \multirow{3}{*}{\textbf{NLLB}} & Pretrained  & 4.54 & 33.53 & 1.17 & 8.23 & 29.51 & 1.03 \\
\multirow{2}{*}{\textbf{NLLB}}          & BPCC  & 5.90 & 36.62 & 1.00 & 9.30 & 33.95 & 0.90 \\
          & Samasāmayik & \textbf{15.83} & \textbf{51.29} & \textbf{0.81} & \textbf{9.51} & \textbf{34.24} & \textbf{0.89} \\
\midrule
% \multirow{3}{*}{\textbf{IndicTrans}} & Pretrained  & 2.39 & 32.09 & 1.11 & 2.39  & 32.09 & 1.11 \\
\multirow{2}{*}{\textbf{IndicTrans}}        & BPCC  & 2.16 & 32.37 & 1.09 & 1.98 & 13.48 & 1.04 \\
        & Samasāmayik  & \textbf{12.25} & \textbf{50.18} & \textbf{0.85} & \textbf{23.49} & \textbf{51.63} & \textbf{0.67} \\
\midrule
% \multirow{3}{*}{\textbf{ByT5}} & Pretrained  & 1.56 & 17.04 & 1.66 & 2.03 & 18.18 & 1.16 \\
\multirow{2}{*}{\textbf{ByT5}}          & BPCC  & 3.34 & 33.99 & 1.34 & 4.59 & 32.29 & 1.26 \\
          & Samasāmayik   & \textbf{10.90} & \textbf{46.53} & \textbf{1.19} & \textbf{20. 05} & \textbf{50.78} & \textbf{0.77} \\
\midrule
\multicolumn{8}{c}{\textbf{IN22}} \\
\midrule
% \multirow{3}{*}{\textbf{NLLB}} & Pretrained  & 2.68 & 28.34 & 1.19 & 13.65 & 39.75 & 0.81 \\
\multirow{2}{*}{\textbf{NLLB}}          & BPCC  & \textbf{9.33} & \textbf{39.25} & \textbf{0.98} & 21.40 & \textbf{45.70} & 0.75 \\
          & Samasāmayik & 2.69 & 31.12 & 1.11 & \textbf{21.41} & 45.69 & 0.75 \\
\midrule
% \multirow{3}{*}{\textbf{IndicTrans}} & Pretrained  & 4.06 & 37.68 & 1.03  & \textbf{16.01} & \textbf{43.11} & \textbf{0.78} \\
\multirow{2}{*}{\textbf{IndicTrans}}   & BPCC  & \textbf{5.80} & \textbf{39.55} & \textbf{1.01} & \textbf{11.75} & 35.35 & 0.84 \\
   & Samasāmayik   & 2.65 & 35.36 & 1.07 & 10.46 & \textbf{36.90} & \textbf{0.83} \\
\midrule
% \multirow{3}{*}{\textbf{ByT5}} & Pretrained  & 0.07 & 7.75 & 1.17 & 0.07 & 9.02 & 1.00 \\
\multirow{2}{*}{\textbf{ByT5}}          & BPCC  & \textbf{7.42} & \textbf{32.26} & \textbf{1.08} & \textbf{14.37} & \textbf{35.88} & \textbf{0.87} \\
          & Samasāmayik & 1.46 & 26.44 & 1.17 & 5.12 & 27.79 & 0.90 \\
\midrule
\multicolumn{8}{c}{\textbf{FLORES (DevTest)}} \\
\midrule
% \multirow{3}{*}{\textbf{NLLB}} & Pretrained  & 1.24 & 28.46 & 1.08 & 15.10 & 40.06 & 0.79 \\
\multirow{2}{*}{\textbf{NLLB}}          & BPCC  & \textbf{1.55} & \textbf{32.02} & \textbf{0.98} & \textbf{17.44} & 43.54 & 0.76 \\
          & Samasāmayik & 1.01 & 30.67 & 1.01 & 17.42 & \textbf{43.58} & 0.76 \\
\midrule
% \multirow{3}{*}{\textbf{IndicTrans}} & Pretrained  & \textbf{2.09} & 32.12 & 1.00 & \textbf{15.43} & \textbf{40.06} &  0.88\\
\multirow{2}{*}{\textbf{IndicTrans}}        & BPCC  & 1.73 & 32.82 & \textbf{0.99} & 1.95 & 13.62 & 0.99 \\
        & Samasāmayik   & 1.38 & \textbf{33.00} & 1.00 & \textbf{11.26} & \textbf{37.38} & \textbf{0.83} \\
\midrule
% \multirow{3}{*}{\textbf{ByT5}} & Pretrained  & 0.08 & 8.83 & 1.12 & 0.06 & 10.35 & 1.00 \\
\multirow{2}{*}{\textbf{ByT5}}          & BPCC  & \textbf{0.95} & \textbf{28.77} & \textbf{1.13} & \textbf{10.06} & \textbf{35.42} & 0.91 \\
          & Samasāmayik  & 0.58 & 28.37 & 1.23 & 6.22 & 30.68 & 0.91 \\
\bottomrule
\end{tabular}
\caption{
Machine translation evaluation results (BLEU, chrF++, and WER) for Hindi$\leftrightarrow$Sanskrit translation using NLLB, IndicTrans2, and ByT5 models. 
}

\label{tab:mt-eval}
\end{table*}

\subsection{Evaluation Data}
\label{sec:evaldata}
In addition to the test data provided as part of Samasāmayik, we use two widely used benchmark dataset for MT evaluation: IN22 and Flores-200.
\paragraph{IN22-Gen}\cite{gala2023indictrans2} is a benchmark for evaluating MT, specifically curated for Indian languages, consisting of n-way parallel data for 22 scheduled languages of India. This dataset is sourced mainly from Wikipedia and other Web sources and the evaluation set consists of 1024 parallel sentences across languages.
\paragraph{Flores-200}\cite{costa2022nllb} is another evaluation dataset widely used as an MT benchmark, which provides parallel data across 200 languages. In this paper, we use the Flores Dev data for Hindi and Sanskrit, containing 997 parallel sentences.

\subsection{Evaluation Metrics}
We use three commonly used metrics for the evaluation of the models for the Machine Translation task including two word-level metrics: Bilingual Evaluation Understudy (BLEU Score) and Word Error Rate (WER) and the character level metric called ChrF++ . The implementation is according to \citet{post-2018-call}\footnote{\url{https://github.com/mjpost/sacrebleu}}. Both the languages Hindi and Sanskrit are morphologically rich languages. Sanskrit can also be agglutinative in nature. Given this property of the languages, we believe that ChrF may be more indicative with its ability to capture the morpho-syntactic aspects of the language.
\paragraph{BLEU Score} is an n-gram precision based metric wherein the n-grams (unigram, bigram, trigram etc.) of the prediction is compared against the respective n-grams in the reference text(s) (ground truth). The computation is as shown below:
\[ BLEU = BP \times \exp{(\sum_{n=1}^{4} w_n \log{p_n})} \]
where BP = Brevity Penalty (to penalise very short predictions), $w_n$ = $\frac{1}{4}$, $p_n$ = fraction of the predicted n-grams that appear in the reference text, 
% $p_n$ - Modified precision means each matching n-gram counts only up to the maximum number of times it appears in a reference.
% The "exp" smoothing method ensures log precision values are non-infinite even when some n-gram counts are 0
\paragraph{ChrF++} \cite{popovic2017chrf++} is a harmonic mean computed over character and word n-gram precision and recall. The computation is shown below:
\[chrF = (1 + \beta^2) \frac{Precison \times Recall}{(\beta^2 \times Precision) + Recall }\]
where, $\beta$~=~weight parameter for Precision and Recall (default value = 2), Precision~=~avg. n-gram overlap precision (prediction $\rightarrow$ reference), Recall~=~avg. n-gram overlap recall (reference $\rightarrow$ prediction).\\
The computation is performed for character n-grams upto n=6 and word n-grams upto n=2.

\paragraph{Word Error Rate (WER)} is derived from the Levenshtein distance, and applied for the MT evaluation at the word level. The definition is as shown in the formula below:
\[ WER = \frac{(S+D+I)}{N}\]
where
S = Number of substitutions, D = Number of deletions, I = Number of insertions, N = Total number of words in the reference text

\subsection{Results and Observation}

\begin{table}[t!]
\centering
\small
\setlength{\tabcolsep}{5pt}
\renewcommand{\arraystretch}{1.2}
\begin{tabular}{lccc}
\toprule
\textbf{Dataset Pair} & \textbf{Jaccard} & \textbf{Cosine} & \textbf{OOV Rate (\%)} \\
\midrule
\multicolumn{4}{l}{\textbf{BPCC vs.}} \\
\hspace{2mm}Flores & 0.196 & 0.600 & 21.52 \\
\hspace{2mm}IN22   & 0.201 & 0.614 & 25.44 \\
\hspace{2mm}Ours   & 0.174 & 0.551 & 55.56 \\
\midrule
\multicolumn{4}{l}{\textbf{Samasamayik vs.}} \\
\hspace{2mm}Flores & 0.174 & 0.551 & 43.93 \\
\hspace{2mm}IN22   & 0.179 & 0.563 & 47.71 \\
\hspace{2mm}Ours   & 0.295 & 0.664 & 18.89 \\
\bottomrule
\end{tabular}
\caption{Lexical similarity and vocabulary overlap between datasets, computed using Jaccard and cosine similarity over token sets, and OOV rate with respect to the reference corpus.}
\label{tab:similarity_metrics}
\end{table}
Table~\ref{tab:mt-eval} shows the comparative results of the models trained on our dataset, against those trained on the existing BPCC data for the task of Hindi-Sanskrit MT. The evaluation results for the models are presented by testing against three test sets: our in-domain test set, IN22 and Flores-200. The results show that the models trained on the training data of Samasāmayik significantly outperform the ones trained on BPCC data when evaluating on the test data of the same. Evaluation on the other 2 datasets have slightly lower performance for Hindi $\rightarrow$ Sanskrit, while the performance is comparable for Sanskrit $\rightarrow$ Hindi.
To measure the significance of the performance differences with different training and evaluation datasets, we compute some similarity metrics between the datasets as shown in the Table~\ref{tab:similarity_metrics}.

Findings: From the evaluation results and similarity metrics, it may be observed that the higher performance gains obtained by the models on certain evaluation datasets may be due to an increased similarity with the training data owing to their common sources. For example, our test set and training set have common sources, while BPCC and IN22 have common sources of data.

\section{Conclusion}
We release a novel dataset named Samasāmayik, a large-scale Hindi-Sanskrit parallel dataset for supporting Machine Translation of contemporary text. The dataset consists of 92,196 parallel sentences from four diverse contemporary domains including spoken tutorials, children's magazine, radio program conversation and language instruction materials, which were carefully curated and aligned by language experts. Our benchmarking experiments with different transformer architectures shows the utility of the dataset for the task of MT and a comprehensive similarity analysis with existing widely used dataset shows its novelty and non-redundancy. By releasing this high-quality, modern prose corpus, we provide a valuable resource to address data scarcity for a low-resource language pair and aim to foster future research in Indic language machine translation.

\section{Limitations}
Although Samasāmayik provides a large-scale contemporary Hindi–Sanskrit parallel corpus, it remains limited to a single language pair and selected contemporary domains. The performance gains observed on in-domain data may not generalise to classical Sanskrit or unseen domains.

\section{Ethics Statement}
All data presented in this paper were collected from publicly available sources and no private or personal data were used. The dataset is released solely for research purposes in Machine Translation and computational linguistics. The corpus includes politically contextualised material (e.g., public speeches) as part of contemporary language usage. Such inclusion is intended purely for linguistic and domain completeness and does not imply endorsement of any political viewpoints, individuals, or institutions.
\section{Acknowledgement}
The authors acknowledge the financial support received from the IKS division, Ministry of Education, Govt. of India through the grant IKS-1891-PHY for the completion of this work. Author N J Karthika was supported by TCS Research Fellowship during her Ph.D. at IIT Bombay. Jahanvi Purohit was supported for Ph.D. by BharatGen.
\bibliography{custom}

\appendix

% \section{Example Appendix}
% \label{sec:appendix}

% This is an appendix.

\end{document}